\documentclass{article}

\usepackage{arxiv} 

	\usepackage[utf8]{inputenc}
	\usepackage{amsmath,amssymb,amsfonts}
	\usepackage{textcomp} 
	\usepackage[british]{babel}	
	\usepackage{csquotes}	

	\usepackage{graphicx}
	\graphicspath{{Figures/}}
	\usepackage[caption=false,font=footnotesize]{subfig}
	\usepackage{dblfloatfix} 
	
	\usepackage{array}
	\usepackage{multirow}
	\usepackage{booktabs}
	\usepackage{adjustbox}
	\usepackage[backend = biber, style=ieee, hyperref = true, url=false, isbn=false, doi=false]{biblatex}
	\DeclareLanguageMapping{british}{british-apa}
	\usepackage{algorithm}
	\usepackage{algorithmicx}
	\usepackage[noend]{algpseudocode}
		\algrenewcommand\algorithmicindent{2.0em}%
		\algnewcommand\Let[2]{\State #1 $\gets$ #2}
		\algnewcommand\AND{\ \textbf{and}\ }
		\algnewcommand\OR{\ \textbf{or} \ }
		\algnewcommand\algorithmicinput{\textbf{Input:}}
		\algnewcommand\Input{\item[\algorithmicinput]}
		\algnewcommand\algorithmiccompute{\textbf{Compute:}}
		\algnewcommand\Compute{\item[\algorithmiccompute]}
		\algnewcommand\algorithmicoutput{\textbf{Output:}}
		\algnewcommand\Output{\item[\algorithmicoutput]}

	\usepackage{enumerate}
	\usepackage{enumitem}
	\usepackage{url}

	\usepackage{color}

	\usepackage[normalem]{ulem}	
	
	\usepackage[]{nohyperref}

	\newcommand{\II}{\ensuremath{\mathbb I}}
	
	\newcommand{\NN}{\ensuremath{\mathbb N}}
	\newcommand{\Argmin}[1]{\ensuremath{\mathrm{Arg}\underset{#1}{\mathrm{min}\,}}}

	\newcommand{\vect}[1]{\boldsymbol{#1}}

	\def\eg{e.g.,\ }

	\renewcommand{\o}{$^{o}$}

	\newcommand{\test}[1]{\ensuremath{{#1}^{\mbox{\scriptsize Test}}}}
	\newcommand{\val}[1]{\ensuremath{{#1}^{\mbox{\scriptsize Val}}}}

\begin{document}

\title{Unsupervised Fault Detection in Varying Operating Conditions\\
{\footnotesize }
\thanks{This research was funded by the Swiss National Science Foundation (SNSF) Grant no. PP00P2\_176878.}
}

\author{Dr. Gabriel Michau\\
\textit{ETH Zurich }\\
Zurich, Switzerland \\
\texttt{gmichau@ethz.ch}
\And
Prof. Olga Fink\\
\textit{ETH Zurich }\\
Zurich, Switzerland \\
\texttt{ofink@ethz.ch}
}

\maketitle

\begin{abstract}
Training data-driven approaches for complex industrial system health monitoring is challenging. When data on faulty conditions are rare or not available, the training has to be performed in a unsupervised manner. 
In addition, when the observation period, used for training, is kept short, to be able to monitor the system in its early life, the training data might not be representative of all the system normal operating conditions. In this paper, we propose five approaches to perform fault detection in such context. Two approaches rely on the data from the unit to be monitored only: the baseline is trained on the early life of the unit. An incremental learning procedure tries to learn new operating conditions as they arise. Three other approaches take advantage of data from other similar units within a fleet. In two cases, units are directly compared to each other with similarity measures, and the data from similar units are combined in the training set. We propose, in the third case, a new deep-learning methodology to perform, first, a feature alignment of different units with an Unsupervised Feature Alignment Network (UFAN). Then, features of both units are combined in the training set of the fault detection neural network.

The approaches are tested on a fleet comprising 112 units, observed over one year of data. All approaches proposed here are an improvement to the baseline, trained with two months of data only. As units in the fleet are found to be very dissimilar, the new architecture UFAN, that aligns units in the feature space, is outperforming others.
\end{abstract}


\section{Introduction}

\subsection{Monitoring Complex Industrial Systems}

With the increased availability of condition monitoring devices, in particular with non-intrusive technologies and retro-fitting, and the decreased costs of data storage, it is now possible to gather large quantities of data on complex industrial systems, a necessity for developing data driven health monitoring approaches, yet, often not sufficient.

Machine learning approaches rely on the assumption that all relevant information on the possible conditions of the system can be inferred from a sufficiently large amount of collected data, referred to as ``training'' data. A consequence of this fundamental assumption is that the performance of such techniques is highly dependent on the representativeness of the training data.

Machine learning approaches have been acclaimed in several fields for their interpolation capability and the very good performances they can achieve on problems where sufficient representative training data is available. 
Since many machine learning tools perform local transformation of the data to achieve better separability, it is very difficult to prove that these transformations are still relevant for data that are outside the value range used for training the models. 
Achieving a robust extrapolation is a largely unsolved problem. If novelty detection methods exist, they face the difficult problem of distinguishing between anomalies and a normal evolution of the system that was not observed in the training dataset.

For the health monitoring of complex industrial systems, these are very strong limitations. Considering that faults are rare (these systems are robust by design and preventive maintenance is usually performed) but potentially numerous, it cannot be expected to collect data on all possible faults, nor to collect enough representative data on faults that did occur. 
Moreover, some industrial systems have very long lifetimes, over which their operation will evolve, due to environment changes, wear and part replacements. In this case, it is also unlikely that a dataset representative of all operating conditions could be collected in a sufficiently short period to start the monitoring of the system in its early life. 
The longer it takes to collect a dataset with representative operating conditions, the longer it will take to provide reliable detection and diagnostics results, and the less advantageous machine learning approaches are compared to traditional ones such as model based monitoring.

In this work, we propose various strategies to apply machine learning to newly installed systems, which rely on the assumption that a limited amount of data have been collected on the system under normal operating conditions. 
Some methods proposed here rely on the assumption that other similar systems have longer and more representative operational experience and that their condition monitoring data are available. 
The later case relies on a fleet of assets and aims at enhancing the representativeness of the training data of a single unit, by transferring knowledge acquired on others. 
It needs to be noted that ``fleet'' can have two definitions: from the operator perspective, a fleet is ``a group of assets organized and operated under the same ownership for a specific purpose''\cite{Jin2015}. The units can vary quite a lot and stem from different manufacturers. The second definition, is from the manufacturer perspective\cite{Leone2017}, where a fleet is a set of units with similar characteristics but that can be operated under different conditions and maintained differently, by one or more operators. 
While in this case, the units might have very different operating conditions, sometimes hardly comparable, similar devices usually monitor the systems.
In this present contribution, we consider this later definition of the fleet. The case study comprises a fleet of 112 gas turbines of a single manufacturer and with similar configurations that are operated in different environments.

\subsection{Related works}

Several previous approaches have been developed for fault detection, diagnostics, prognostics and maintenance for fleets of systems, with the focus on transferring information between units of a fleet~\cite{AbhinavSaxena2014,Jin2015a,Al-Dahidi2016,Gonzalez-PriDa2016,Liu2016,Leone2017,Peysson2019}.
Most of the approaches have been developed for a supervised set-up where the main challenge is learning and transferring the relevant fault specific patterns or degradation behaviour between units.
Some works have focused on inferring models or statistical features relevant at the fleet level and to fine tune them for each unit~\cite{Peysson2019,Liu2016,Zhang2017a}. Such approaches are particularly powerful when the assets and their operating conditions are sufficiently similar so that small modifications of the features are sufficient to achieve the transfer between units.
In fleets with more variations in characteristics and operating conditions, an identification of similar units in sub-fleets is required, such as for example the diversity indices proposed in~\cite{Gonzalez-PriDa2016} based on operating profiles. 
The main challenge for fleets of dissimilar units is to identify the right indices that will group sub-fleets with relevant similar characteristics. 
This, typically, requires a lot of manual work and expert knowledge on the systems and their operating conditions.

Another approach to identify similar units is to directly compare the distances between datasets, a problem affected by the curse of dimensionality: in high dimensions, the notion of distance loses its traditional meaning~\cite{Domingos2012}, and the temporal dimensions particularly important when operating conditions evolve, make this comparison even more challenging. 
Previous works have mostly focused on comparing one-dimensional time series~\cite{Leone2016} or comparing multi-dimensional measurements, independent of time~\cite{Zio2010}. 
In~\cite{Lapira2012}, the fleets clustered based on single dimensional time series representative of the specific operating conditions (\eg wind speed for a wind turbine fleet). Subsequently, the measurements are clustered in sub-fleets for fault detection.

When sufficiently similar units do not exist or cannot be identified, domain adaptation has recently been used to transfer knowledge between units or between different operating conditions within the same unit. Again, most approaches have been developed for supervised set-ups. In such a framework, the unit for which the application is developed is denoted by ``target'' while the datasets used for enhancing the training is denoted as ``source''. In~\cite{Li2018}, a two level deep learning approach is used, first, to generate fake samples in different operating conditions and second, to train a classifier also on operating conditions where the faults were not experienced. In~\cite{Xie2016}, \citeauthor{Xie2016} proposed feature extraction and fusion using transfer component analysis. In \cite{Lu2017a}, a neural network is trained such as the latent space minimises the Maximum Mean Discrepancy~\cite{Borgwardt2006} between the features of the source and the target while maximising the detection accuracy of a fault classifier. Surprisingly, very few works have applied similar approaches to fleets of assets.

In the present contribution, we build upon our previous work on unsupervised fault detection with Hierarchical Machine Learning~\cite{Michau2017,Michau2018a,Michau2018b,Michau2018c}, to address the problem of unsupervised detection, in the context when only little condition monitoring data are available. 
We focus on two strategies: first, we consider models that are robust against the evolution of the operating conditions of a unit, based on the data of the unit of interest only; second we consider models whose training set is enhanced with data from other similar units. In the first case, the challenge lies in being robust to evolutions while also being able to detect anomalies.
In the second case, the difficulty is to find data that are relevant to the unit of interest: if data are too similar, they might be redundant only, if they are too dissimilar, the ability to detect faults might not be retained.

In this paper, we explore both strategies with six different approaches. First and second, two naive baselines are used where an Hierarchical Extreme Learning Machine (HELM) is trained with condition monitoring data from a single unit. One is trained with all available data (for comparison purposes only), the other is trained on the early life of the unit only.  
Third, an incremental learning procedure is proposed where data from the unit are tested in small batches and added to the training set when the number of alarm raised is sufficiently low. Fourth and fifth, pairs of units are compared with two distance measures, one based on the HELM magnification coefficient, the other based on the Maximum Mean Discrepancy (MMD). The data of the pair minimising these distances are combined in a single training set and used for training a HELM. Last, an Unsupervised Feature Alignment Network (UFAN) is proposed, such as to align data from different units in the feature space. Features from both units are then combined to train an unsupervised fault detection network. 

\subsection{Stator Vane Case Study}

To demonstrate the suitability and effectiveness of the proposed approaches and compare between the different strategies, a real application case study is used that faces the difficulties discussed above, including rare faults, limited observation time, limited representativeness of condition monitoring data collected over a short periods.																																				
The comparisons are performed on a fleet comprising 112 power plants. 
The case study is identical to that presented in ~\cite{Michau2018b}. 
In the available fleet, about 100 gas turbines have not experienced identifiable faults during the observation period (approximatively one year) and 12 units have experienced a failure of the stator vane. 

A vane in a compressor redirects the gas between the blade rows, leading to an increase in pressure and temperature.
The failure of a compressor vane in a gas turbine is usually due to a Foreign Object Damage (FOD) caused by a part loosening and travelling downstream, affecting subsequent compressor parts, the combustor or the turbine itself.
Fatigue and impact from surge can also affect the vane geometry and shape and lead to this failure mode. Parts are stressed to their limits to achieve high operational efficiency with complex cooling schemes to avoid their melting, especially during high load.
Such failures are undesirable due to the associated costs, including repair costs and operational costs of the unexpected power plant shutdown. 

Because of the various different factors that can contribute to the source of the failure mode, including assembly, material errors, or the result of specific operation profiles, the occurrence of a specific failure mode is considered as being random. Therefore, the focus is nowadays on early detection and fault isolation and not on prediction.

So far, the detection of compressor vane failures mainly relied on analytic stemming from domain expertise. 
Yet, if the algorithms are particularly tuned for high detection rates, they often generate too many false alarms. 
False alarms are very costly, each raised alarm is manually verified by an expert which makes it a time- and resource-consuming task.

The data available in this case study is limited to one year, over which the gas turbines have not experienced all relevant operating conditions. We aim at being able to propose condition monitoring methods that rely on two months of data only from the turbine of interest.

The remainder of the paper is organised as follows: 
Section 2 gives a recap on the theory of hierarchical extreme learning machines and motivates their usage in the particular context of unsupervised learning. 
Section 3 proposes a new method for feature learning that relies on adversarial learning in a fully unsupervised set-up. 
Last, Section 4 presents the case study, the different models tested and compared, and discusses the results that could be achieved.

\section{Hierarchical Extreme Learning Machines for Unsupervised Fault Detection}

\subsection{Motivations}

Extreme Learning Machines are single layer feed-forward networks. 
Their distinguishing characteristic is that the weights and biases between the input layer and the hidden layer are drawn randomly, while only the weights between the hidden layer and the output are learned. 
These networks are universal approximators, they can approximate any function with any accuracy~\cite{huang_universal_2006}.
Their advantage is the simplicity and rapidity of their training which consists in solving a single variable convex optimisation problem. 
The iterative and time consuming back-propagation is therefore not required.

Deeper architectures are achieved with ELM by stacking networks on top of each other. 
These are the so-called Hierarchical Extreme Learning Machines (HELM). 
This stacked approach consists in extracting for each ELM its hidden layer and to use it as input to the next. 
As no back-propagation can be used, the information is flowing forward only, this approach requires the sequential training of each ELM. 
The lower networks are usually trained in an unsupervised manner, most of the time as auto-encoders, while only the last network is trained for the main task (regression, classification, anomaly detection). 
This architecture has the advantage of mimicking more traditional approaches which usually require the extraction of statistically (or physically) significant features that will be used in a second step for the task of interest. 
With HELM, the auto-encoders are, in fact, feature learners.

In the context of complex system monitoring, faults are rare and lack representative data, because of preventive maintenance or because the fault to come has never yet occurred. 
It is therefore paramount to propose detection methods that do not aim at identifying past faults but can also detect new abnormal operating conditions. 
Previous works have demonstrated and analysed the performance and the merits of HELM architectures for unsupervised fault detection~\cite{Michau2018a} and fault isolation~\cite{Michau2017} compared to other machine learning approaches. 
HELM with a single compressive auto-encoder ELM stacked with a one-class classifier, trained on healthy data only, has given very encouraging results.
In fact, under the assumption that healthy data only is available for the training, it is not possible to quantify during the training, how well the features can help to discriminate between healthy and unhealthy conditions. 
It is therefore not desirable to back-propagate the loss of the one-class classifier to the feature learner, as it conveys no information on the ability of the features to discriminate faults. This justifies \textit{a priori} the suitability of ELM-based architectures. In \cite{Michau2018a,Michau2017}, it has been found that the output of the one-class classifier ELM can be interpreted as a distance to the dataset used in the training of the network. A threshold on this distance can then be used to distinguish points belonging to the main class (the one used for the training, that is, the healthy class) and outliers. Moreover, the value of this distance could also be interpreted as the severity of the anomaly. Once a fault has been detected, the analysis of the residuals in the auto-encoder of the HELM could provide with high accuracy the signals responsible for the anomaly detection~\cite{Michau2017}.

In these previous works, the necessity of the feature learning steps for better accuracies has been demonstrated and justifies the use of HELM compared to a simpler one-class classifier ELM alone. In this work, we therefore build on this knowledge and will only consider such hierarchical architectures, either HELM with a single auto-encoder stacked with a one class classifier, or an ELM one-class classifier on top of another feature learner.

\subsection{ELM-based models}

\subsubsection{Principle}
The single layer feed forward network equation is
\begin{equation}
\label{eqn:elm}
Y = g(\vect{A}, X, B)\cdot \vect{\beta},
\end{equation}
where $X$ in the input, $Y$ the output, $g$ an activation function, $\vect{A}$ and $B$ the weights and biases connecting the input layer to the hidden layer and $\beta$ the weights connecting the hidden layer to the output.

The ELM training consists in first drawing $\vect{A}$ and $B$ randomly, and then in solving Eq.\eqref{eqn:elm} such as $Y$ is as close as possible from a desired target $T$. The quality of the solutions of such inverse problems are known to depend on a regularisation function, which can prevent overfitting and insure parsimony. Finally, given an architecture, $\vect{A}$, $B$ and a target $T$, the optimal ELM is found for $\beta$ satisfying
\begin{equation}
\label{eqn:beta_est}
\vect{\hat{\beta}} = \Argmin{\vect{\beta}} \Vert g(\vect{A}, X, B) \vect{\beta} - T \Vert_u^{\sigma_1} + C \Vert \vect{\beta} \Vert_v^{\sigma_2 }
\end{equation}
where $\sigma_1, \sigma_2 >0$, $u, v\in \NN\backslash 0$ and $C\geq 0$ is the weight of the regularisation.
$C$ represents a compromise between the learning objective and the regularisation of $\vect{\beta}$ that one would like to impose (\eg sparsity, non-diverging coefficient, etc...).

\subsubsection{Auto-encoder}

An auto-encoder is a machine learning approach aiming at performing a transformation of the inputs in a latent space and then to find the best inverse transformation to reconstruct the input. In many applications, the latent space is taken of smaller dimensionality than the input, either for data compression or to avoid overfitting, by excluding dimensions with lower explanatory power. These are denoted by compressive auto-encoders.

For the auto-encoder ELM, we proposed in~\cite{Michau2018a} to use a compressive auto-encoder and to enforce sparsity in the coefficient $\beta$ such as to minimise the number of input signals linked to each feature. This aims at increasing feature specificity in part of the system.
For the ELM auto-encoder, Eq.\eqref{eqn:beta_est} becomes
\begin{equation}
\label{eqn:beta_est_l1}
\vect{\hat{\beta}} = \Argmin{\vect{\beta}} \Vert \vect{H}\vect{\beta} - X \Vert_2^{2} + \lambda \Vert \vect{\beta} \Vert_1.
\end{equation}
The equation is solved with the FISTA algorithm \cite{beck_fast_2009,Michau2018a}.

\subsubsection{One-class classifier}

One-class classification aims at learning the characteristics of a single class, to then estimates if new data points belongs to that class or are outliers. Most of the one-class classification methods end up monitoring a distance to the training cloud of points and to learn a boundary beyond which points are considered as outliers. This is a threshold-based decision.
In \cite{Michau2018a}, we demonstrated that ELM trained as a one-class classifier is competitive with other machine learning approach, including the famous one-class SVM.

For the one-class classifier ELM, we assume that its input will be a set of parsimonious and relevant features. Thus, we only impose a $\ell_2$ regularisation in Eq.\eqref{eqn:beta_est}, mostly to avoid exploding coefficients. Eq.\eqref{eqn:beta_est} becomes a Ridge regression problem whose solution has a closed form:
\begin{equation}
\label{eqn:ridge_sol}
\vect{\beta} = \left( C\cdot\II + \vect{H}^\top\vect{H}\right)^{-1}\vect{H}^\top T
\end{equation}

In \cite{Michau2018a}, we demonstrated that a good decision boundary on the output of the one-class ELM is the threshold
\begin{equation}
\label{eq:thrd}
\mbox{Thrd} = \gamma\cdot \mbox{percentile}_{p}(\vert 1 - \val{Y}\vert),
\end{equation}
with $\gamma \in [1,2]$ and $p\in [95,100]$.
Outliers are discriminated from the main healthy class if they are above this threshold. In addition, the severity of the anomaly can be assessed with the ``magnification coefficient'' defined as
\begin{equation}
\label{eq:mag}
\mbox{Mag} = \frac{(\vert 1 - \test{Y}\vert)}{\mbox{Thrd}}.
\end{equation}

\subsubsection{HELM}
\begin{figure}[htpb]
\centering
\includegraphics[width=7cm]{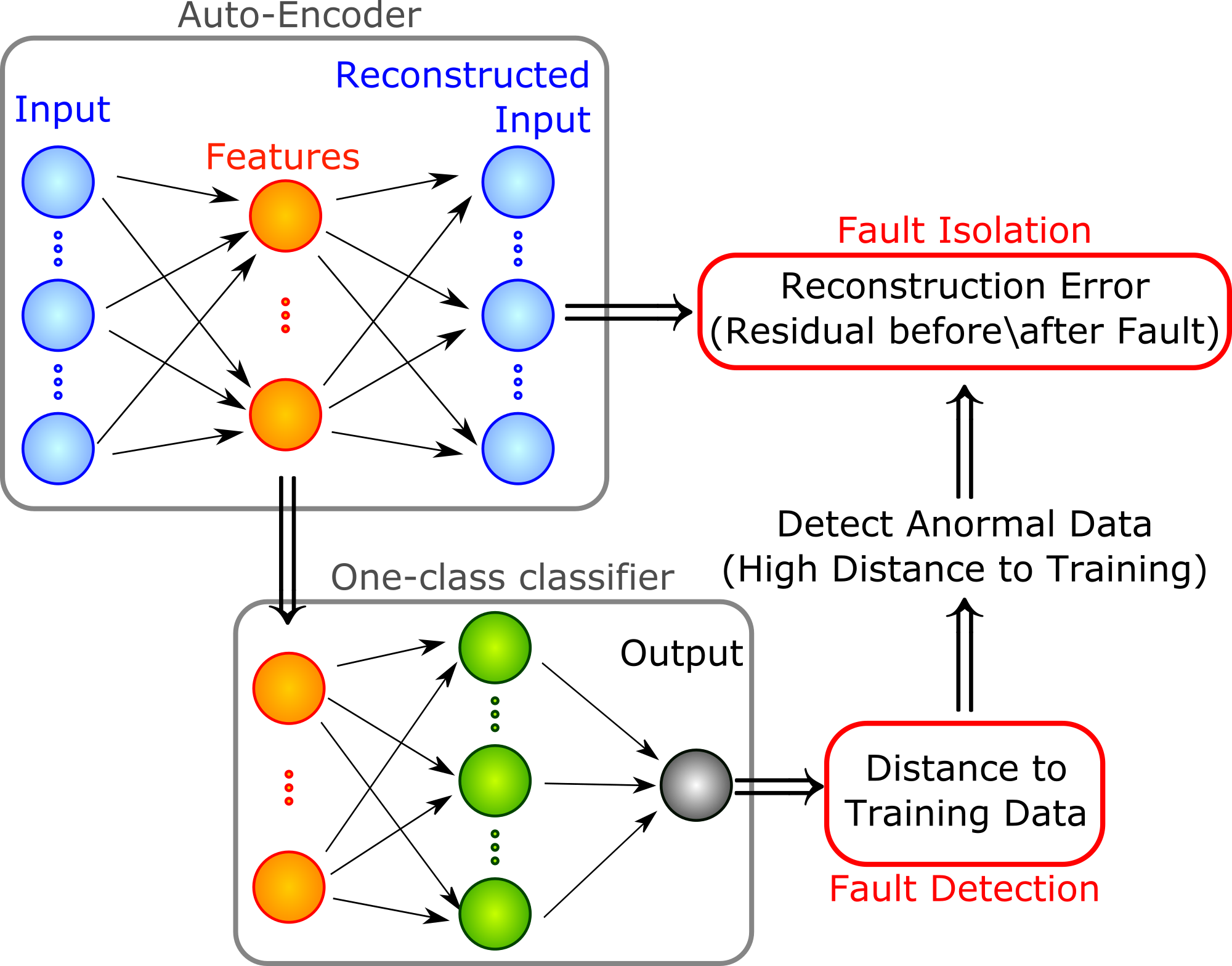}
\caption{\textbf{HELM architecture}. The HELM consists of an arbitrary number of stacked unsupervised auto-encoder ELM and of one last layer trained for the task at hand. In our case it is a one-class classifier ELM.}
\label{fig:HELM}
\end{figure}
The HELM architecture used in this work is presented in Figure~\ref{fig:HELM}. 
A single compressive ELM auto-encoder is used as a feature learner. 
These features are then used as input to the one-class classifier ELM. 
Using the decision boundary described in Eq.~\eqref{eq:thrd}, abnormal data points can be detected. 
The fault isolation can be performed based on the residuals of the auto-encoder at detection time, such as presented in ~\cite{Michau2017}. Yet, fault isolation is not the focus of this contribution.

\section{Feature Alignment \& Cross Domain Adaptation}

\subsection{Context}
Inspired by the latest advances in adversarial training, deep learning recently had many applications in cross-domain adaptation, mainly in the field of image processing. 
Cross-domain adaptation consists in combining information from similar datasets, usually with more data available, to improve results on classic machine learning tasks. 
In many applications, the approach aims at enhancing the training set, with data acquired in a different set-up.
For images, this could be pictures acquired with different devices (camera, webcam, etc...) or in different context (\eg numbers on different background or with different textures)~\cite{Patel2015}. 
In the PHM field, this could be data from different operating conditions or from other units~\cite{Li2018}.

Very promising results have recently been achieved with adversarial architectures, combining a feature learner, a discriminator and last a network performing the task of interest (mostly regression of classification)~\cite{Ganin2016}, denoted in the following, without loss of generality, by ``main classifier''. 
The feature learner aims at encoding the raw data from both datasets, source and target, to a latent space $F$. 
Then, the data encoded in this latent space are used as input to the two other networks, the discriminator and the main classifier. 
The discriminator is a classifier trained to classify the origin of the data (source or target).
The main classifier is trained in a supervised manner with the features of the source. 
The adversarial training is performed at the level of the feature learner, trained as to minimise the main classifier loss and maximise the discriminator loss. 
The learned features should at the same time contain the information necessary for the main classification task while being independent of the dataset of origin. 
This process is also referred to as distribution alignment~\cite{MariaCarlucci2017}.

The training of such architecture has been simplified by the introduction of the Gradient Reversal Layer trick~\cite{Ganin2016}. 
This non trainable layer, added in between the feature encoder and the discriminator is designed to apply the identity function in the forward pass, and to flip the gradient in the backward pass (it applies a negative factor to it). 
The standard optimisation technique is performing a gradient ascend on the weights of the first network with respect to the discriminator loss, and thus fosters its maximisation.

\subsection{Unsupervised Feature Alignment Network (UFAN)}
In the context of fault detection for units under evolving operating conditions, such adversarial architectures could be used to align healthy operating conditions from different units and to combine the different operating experiences in a single model. 
This aims at making the fault detection approach more robust to future evolution of the source unit operation. 
One issue preventing the direct application of other works found in the literature to our context is that, in our case, the classification task is done in a complete unsupervised manner (with respect to the faults to be detected), with healthy data only. 
As discussed previously, it is not desirable to back-propagate the loss of the one-class classifier. 
Instead, we propose a new loss for the adversarial training, which aims at ensuring that the features are representative of the input data. 
This loss measures the distance to the closest homothetic transformation between the input space and the feature space, and aims at ensuring the conservation of the relative distances between pairs of data-points in the feature space. 
This loss allows for the adversarial training of the feature encoder in an fully unsupervised manner. 
As back-propagation from the one-class classifier to the rest of the architecture is not desired, and for consistency with the other approaches discussed here, we will also use a one-class classifier ELM as in other models.

\begin{figure}[htbp]
\centering
\includegraphics[width=7cm]{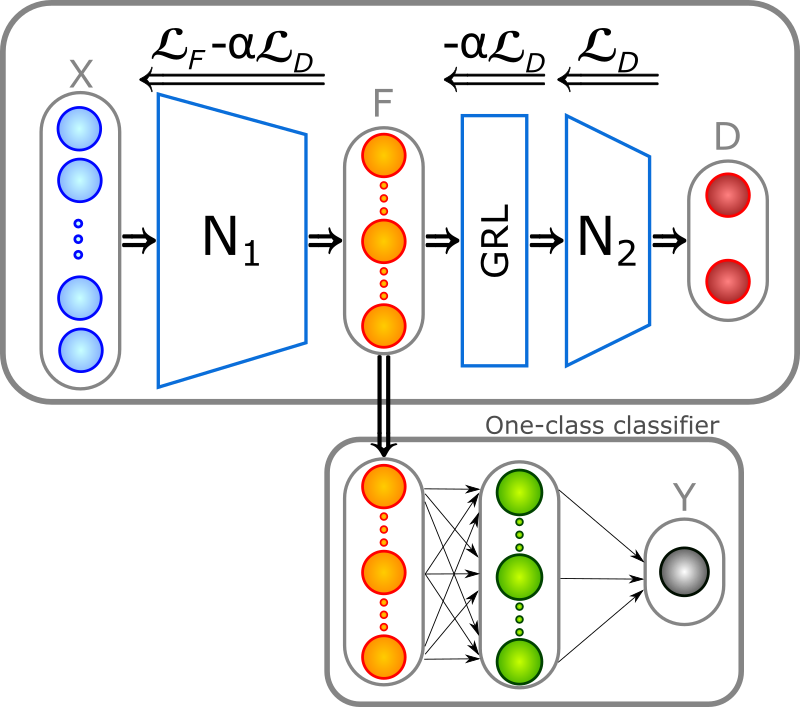}
\caption{\textbf{UFAN Architecture}. 
Our Unsupervised Feature Alignment Network consists of three networks. A feature encoder $N_1$, a discriminator $N_2$ and a one-class classifier ELM. 
The discriminator is trained to minimise the classification loss $\mathcal{L}_D$ on the origin of the data (source vs. target) and is connected to the feature encoder with a gradient reversal layer (with factor $-\alpha$ on the gradient). 
The feature encoder is trained as to minimise the loss on the homothety between $X$ and $F$, $\mathcal{L}_F$, but as to maximise the loss of the discriminator $\alpha\mathcal{L}_D$, so that the origin of $X$ in $F$ is undistinguishable.}
\label{fig:UFAN}
\end{figure}

The proposed architecture is illustrated in Figure~\ref{fig:UFAN}. 
The two losses used for the Unsupervised Feature Alignment Network, made of the feature encoder $N_1$ and the discriminator $N_2$ are, the binary cross-entropy for the discriminator loss $\mathcal{L}_D$ and the homothetic conservation loss $\mathcal{L}_F$. 
$\mathcal{L}_F$ is defined as:

\begin{equation}
\mathcal{L}_F = \sum_{S\in \left\lbrace \substack{\mbox{Source}\\\mbox{Target}} \right\rbrace} \frac{1}{\vert S \vert}  \sum_{(i,j)\in S}\left\Vert \left\Vert X_i - X_j \right\Vert_2 - \eta \left\Vert F_i - F_j \right\Vert_2 \right\Vert_2
\end{equation}
where
\begin{equation}
\eta = \Argmin{\tilde{\eta}} \mathcal{L}_F (\tilde{\eta})
\end{equation}

\section{The case Study}

\subsection{Presentation of the data}
The case study comprises a fleet of 112 gas turbines, 12 of which experienced a stator vane fault. 
For the 100 remaining ones, the experts in the available data could not identify any particular fault.
We will therefore assume that their data are healthy. 
They compose a fleet of healthy unit, each with one year of data available. 
We will test our approaches on the 12 units which experienced a fault, and compare the models based on their ability to detect the fault while limiting the number of false positives. 
To do so, we assume that the first two months of data are available for the training. 
All data points after these two months and until a month before the expert detected the fault, are considered as healthy, and are used to quantify the number of false positives. 
The last month before the detection is ignored as faults could be the consequence of prior deteriorations and a detection could not be reliably compared to any ground truth. 
Last, the data points after the time of detection by the expert are considered as unhealthy. 
As many of the 12 datasets have very few points available after that detection time, we will consider the fault as detected if the threshold is exceeded at least once.

The measurements are sampled every 5 minutes over 1 year. They stem from 15 real sensors 9 ISO variables (measurements modified by a physical model to represent some hand-crafted features in standard operating conditions 15\o C, 1 atmosphere).
Available ISO measurements are, the power, the heat rate, the efficiency and indicators on the compressor (efficiency, pressure ratio, discharge pressure, discharge temperature, flow).
Other measurements are pressures and temperatures from the different parts of the turbine and of the compressor (inlet and Bell Mouth), ambient condition measurements and operating state measurements such as the rotor speed, the turbine output, and fuel stroke. 

Two months of data represents around 17\,000 samples per measurement. 
The false positives are evaluated over around 9 months or 39\,000 samples. The true positives are available for at least few hours and their number depends on the unit (from 10 to 1000 points). 

The data has been normalised such that all variables 1st and 99th-percentiles are respectively $-1$ and $1$, and rows with any missing values or 0 (which is not a possible value for any of the measurement) have been removed. 
Every data used in the training with values above $3$ after normalisation have been removed.

The main differences with our previous study in~\cite{Michau2018b}, are that more models are compared to each others, in particular the new architecture UFAN. Also, all variables are used as input of the models, not ISO only as in~\cite{Michau2018b}.

\subsection{Model architecture and hyperparameters}

For this case study, we propose to compare two different approaches with six models. First, three models that are trained on a single unit only. The baseline models are an HELM trained with respectively nine and two months of data only. The third model is an HELM trained incrementally every two weeks. Second, three models use another units from the available fleet to extend the training. As these models perform a pairwise association of the target with another source unit, they are denoted in the following by ``pairwise models''. Two of the models rely on two different similarity measures and uses HELM as a monitoring method, the third one relies on the UFAN presented above.

All the feature extractor used in this paper (auto-encoder ELM and the UFAN) have 10 features. The ELM auto-encoder have a LASSO regularisation factor of $10^{-3}$.
All one-class classifier ELM have 200 neurons and a Ridge regularisation factor of $10^{-5}$. All ELM-based results are averaged over 8 models as proposed in ~\cite{Michau2018a}. ELM networks training is very efficient and even with 8 models, their memory requirements and computational costs remain negligible against all other tools used in this contribution. Computation times are provided with the results.

The detection of abnormal data points with the one class classifier is performed according to Eq.~\eqref{eq:thrd} with $p=99.5$ and $\gamma=1.2$.
These hyperparameters are in line with the best sets identified in~\cite{Michau2018a, Michau2018b}.

\subsubsection{Single Unit based Models}

Three models are trained using the data of the single unit only. 
The first model, used as a very naive baseline is an HELM trained on the first nine months of data. 
It mimics the situation were an operator would be ready to wait almost one year before to start monitoring the unit, in the hope to collect many different operating conditions.

The second model is a more realistic baseline, where an HELM is trained on the first two months of data and tested as described above. 
No further hyper parameter needs to be specified.

The third single unit model proposes a naive approach to incremental learning. 
An HELM is trained on the first two months of data of the single unit. 
Then, it is tested on the next two weeks of data. 
If the number of false positives in these two weeks is lower than a ratio $r$, the two weeks are added to the training set and the HELM is re-trained, otherwise the two weeks are ignored. 
This process is repeated on all data available for the unit.

This approach has been tested for different $r$ and only the best results have been reported here. 
It has also been tested with a sliding windows of two months for the training, where each time two weeks are added to the training set, the oldest two weeks are removed. 
These two approaches gave very similar results and the later one will not be detailed further.

\subsubsection{Pairwise Association of Assets}

Three approaches are tested where the unit of interest, the target, is compared to all other in the fleet of 100 healthy units. When a suitable candidate is identified, the data of both units (the two months of the target and the one year of the source) are used for training the models.

As proposed in~\cite{Michau2018b}, our first approach is to use an HELM to measure dataset similarities. 
In fact, the HELM output is similar to a distance to the training class and can therefore be used to compare two datasets to each other. 
To do so, an HELM is trained on the two months of data of the target unit and tested on all units of the fleet. The units with lowest average magnification coefficient are considered as the closest units, and as candidate source units.

In a second steps, using the source and target data altogether, a new HELM is trained and is used to monitor the target unit.

The second approach is very similar, but instead of comparing the sets with an HELM, the Maximum Mean Discrepancy (MMD) is used instead~\cite{Borgwardt2006}. We faced the problem of memory issues when computing the MMD over several tens of thousands of data points. We mitigated this by splitting the datasets in chunk of one month.
The source unit is selected among the fleet of healthy ones, as the one that minimises the average MMD over the year. Then, similarly as above, an HELM is trained with all data points available from both source and target to monitor the target.

For the MMD, we considered a set of 6 Gaussian kernels with width $0.05$, $0.1$, $0.2$, $0.5$, $1$ and $2$ respectively.

The last approach consisted in training the UFAN (cf Fig.\ref{fig:UFAN}) with the target and every candidate source unit. The selected candidate would be the one minimising the final adversarial loss $\mathcal{L}_F-\alpha \mathcal{L}_D$. This would be the unit whose features could best be aligned, while confusing the discriminator. The one-class classifier ELM is then trained with the features of both the source and the target.
In the UFAN, the feature encoder comprises two fully connected layers of 15 and 10 neurons respectively. The discriminator is also made of two fully connected layers, of 10 and 5 neurons respectively. The training is performed in batch of 150 samples over 100 epochs with a learning rate of $10^{-4}$. Different values for $\alpha$ have been tested ($0.1$, $1$, and $10$), yet as results where not radically different, only the case $\alpha=1$ is discussed further.

\subsection{Results}

\subsubsection{Results presentation}
Table~\ref{tbl:results} presents the results achieved by the six models for the 12 units of interest.
The six models are denoted by a key: 
`H' for HELM based model and `UFA' for the UFAN architecture. `H-9m' and `H-2m' are the HELM trained with respectively nine and two months of data, `H-Inc' is the incremental learning model, `H-H' is the pairwise model with the source unit minimising the average magnification coefficient of an HELM trained on the target and `H-M' is the pairwise model with the source unit minimising the MMD distance with the target. 
Results for the pairwise models in the first part of the table (columns 5 to 7) are only presented for the candidates minimising the selection criteria as described above.
In this table, we consider a model as ``valid'' if it detects the fault (otherwise results are not reported) and if the number of false positives is below 15\%. 
This is justified by the fact that detecting the fault is a requirement and that models with a high false positives rate will never be considered for implementation. 
The two bottom lines of the table contain the number of valid models over the twelve units, and the average of false positive percentages over the valid models only.

From the first half of Table~\ref{tbl:results} (up to column 7), it can be verified that all approaches are an improvement to the baseline training of two  months (3rd column). 
The UFAN approach leads to valid model for ten of the twelve units. 
The two missed units are the most challenging ones and were missed by all models but the ones trained on the units selected based on the MMD distance (units 2 and 10).

When HELM is trained on 9 months of data, it is the second best approach which proves that even when gathering almost one year of data, not all relevant operating conditions might have been observed and that enlarging the training set with other units is still a relevant approach.

The models trained on pairs selected based on their MMD distance lead to valid models in seven of the twelve units, which is in line with the incremental leaning approach. 
The MMD distance can lead to valid models where other approaches fail. 
However, it fails in cases where the other approaches do not seem to be challenged. 
The embedding in a kernel space modifies the meaning of similarity and it could help in some cases to identify relevant pairs that would be missed by other approaches. 

The incremental learning approach appears to be like a black \& white approach. 
When the model is valid, it performs very well with almost no false positives at all, while for other models it has very high false positives detection rates. 
This is probably due to the large ratio $r$ used for deciding whether the next two weeks of data should be added to the training. 
Lower $r$ reduces drastically the number of valid models (the models learn less and less as more and more weeks with data are excluded from the updates) and results become very close to the model trained with two months of data only. 
Then the pair of units selected with HELM provides results comparable to the MMD-based selection. 
The results are also in line with those from~\cite{Michau2018b} that highlighted the difficulty, for this case study, to identify similar units.
This difficulty explains and proves the benefits of imposing feature alignment. 
When units are dissimilar, aligning the features is necessary to benefit from other units.

The second part of the table, presents results over all possible pairs, for both HELM and the UFAN model. 
The first two columns of the left part (columns 8 and 9) present the best results achieved over all pairs in both cases. 
These results are better than those achieved with the selection criteria proposed here. 
It looks therefore like the distances between units, at least the ones proposed here, are not the optimal selection criteria. 
This remark is supported by the weak correlation between the distance between units and the performance of the models. The number of false positives of the models is moderately correlated with the average HELM magnification coefficient (0.45), and weakly correlated with MMD (0.3) and the final UFAN loss (0.2). 
It has to be noted that if a better selection criteria could be found, the results of the UFAN approach would be close to ideal, with all faults detected and very little false positives (maximum of 1.3\% on the most challenging unit 12).

\subsubsection{Beyond the results}
\begin{figure*}
\centerline{\includegraphics[width=14cm]{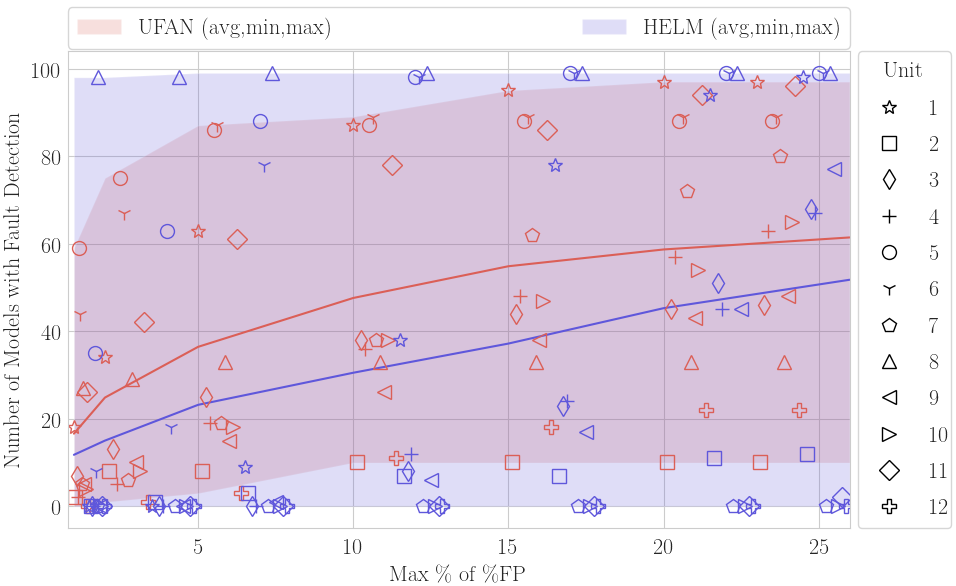}} 
\caption{\textbf{HELM versus UFAN} Number of source units, which once paired with the target led to a valid model, that is: the fault has been detected and the percentage of false positive is smaller than the value in abscissa. The continuous line is the average, the region is the min and the max over the twelve individual units and each of the twelve pattern represents one unit.}
\label{fig:HELMvsGRL}
\end{figure*}

As the proposed distances are not optimal selection criteria, we propose a deeper analysis of results over all possible pairs. 
The last four columns of Table~\ref{tbl:results} present the number of valid models that could be achieved over the 100 pairs and the average false positives percentage. 
These results are also plotted in Figure~\ref{fig:HELMvsGRL}. 
In Figure~\ref{fig:HELMvsGRL}, the number of valid models is plotted against the threshold under which a model is considered as valid (from 1\% to 25\% of false positives allowed). 
The continuous line is the average over the twelve units and the regions enclose the minimum and the maximum. Individual results per unit are also plotted with patterns, for information only.

The results found here are in line with \cite{Michau2018b} where similar units could be identified for less than half of the fleet. Aligning the features with the UFAN approach match more units together and is a real improvement. It finds not only at least one matching unit for five sixth of the fleet, but it matches in average half of the fleet to any unit. In addition, the final resulting false positive rates are much lower.

Last, as expected, running dataset comparison in multi-dimension based on distances is a very expensive process. We ran the experiments with Euler V\footnote{\url{https://scicomp.ethz.ch/wiki/Euler}}, with two cores from a processor Intel Xeon E5-2697v2. For an average pair of units, computing the MMD required 80GB of RAM and took in average 650 seconds, training and running HELM with 8 repetitions required 1GB of RAM and took 7 seconds, and training an running UFAN needed 1GB of RAM and took 90 seconds. Given the cost of MMD compared to HELM with little improvements in the results, it confirms that HELM used as a distance measure is relevant and efficient. 
The UFAN approach is more costly than HELM but with high benefits on the results.

\begin{table}
\setlength\tabcolsep{3pt}
\caption{Results of the six Models Achieved on the 12 Units.}
\begin{tabular}{l|r|rr|rrr||rr|rr|rr}
\toprule
\           & \multicolumn{3}{c|}{Single Unit Training}&\multicolumn{3}{c||}{Pairwise Models}& \multicolumn{2}{c|}{Best Model} & \multicolumn{2}{c|}{\# valid pairs} & \multicolumn{2}{c}{Mean \%FP}\\
\cmidrule{2-4}\cmidrule{5-7}\cmidrule{8-9}\cmidrule{10-11}\cmidrule{12-13}
Unit        & H-9m & H-2m & H-Inc     & H-H & H-M     & UFA           & HELM & UFA  & HELM                & UFA & HELM & UFA \\
\midrule
1         & 0  & 9.0     & {0.0} & 8.9       & 8.5          & 3.7           & 2.2  & 0.0  & 78                  & 95  & 10.0   & 4.4    \\
2         & \  & \       & \            & \         & 5.2          & \             & 1.4  & 0.1  & 7                   & 10  & 7.0    & 2.3    \\
3         & 42.3  & 18.9    & {0.0} & 14.9      & \            & 4.9           & 5.7  & 0.2  & 23                  & 44  & 12.2   & 5.3    \\
4         & 0.3  & 42.6    & {0.0} & 7.8       & 30.8         & 10.7          & 5.7  & 0.6  & 24                  & 48  & 10.6   & 7.1    \\
5         & 3.7  & 4.2     & {0.0} & 0.7       & 0.1          & 0.4           & 0.0  & 0.0  & 99                  & 88  & 2.1    & 1.1    \\
6         & 2.5  & 3.1     & {0.0} & 5.9       & 0.6          & 0.5           & 0.5  & 0.0  & 99                  & 89  & 3.7    & 1.4    \\
7         & 4.3  & 56.5    & 86.5         & 42.3      & 15.4         & {15.3} & 28.6 & 0.0  & 0                   & 62  & \      & 8.0    \\
8         & 0.4  & 0.2     & {0.0} & 0.2       & \            & 0.2           & 0.0  & 0.0  & 99                  & 33  & 0.2    & 0.5    \\
9         & 12.5  & 20.1    & {0.0} & 17.2      & \            & 5.4           & 4.8  & 0.0  & 17                  & 38  & 11.9   & 6.8    \\
10        & 96.2  & 81.8    & 73.3         & 76.8      & {5.0} & 32.2          & 29.2 & 0.1  & 0                   & 47  & \      & 6.4    \\
11        & 0.8  & 36.2    & 82.9         & 48.7      & 2.1          & {1.1}  & 20.6 & 0.0  & 0                   & 86  & \      & 3.9    \\
12        & 100  & 68.6    & 89.7         & 53.1      & \            & {6.0}  & 35.3 & 1.3  & 0                   & 18  & \      & 9.3    \\
\midrule
\# Val. & 8   & 4       & 7            & 6         & 7            & {10}   & 8    & 12   & \                   & \   & \      & \      \\
Mean      & 3.1 & 4.1     & 0.0          & 6.4       & 5.3          & 4.8           & 2.6  & 0.2  & 37                  & 55  & 7.2    & 4.1    \\
\bottomrule
\multicolumn{13}{p{11cm}}{\textbf{Abbreviations}: `H-9m' and 'H-2m' for HELM trained on 9 and 2 months of data. `H-Inc' for incremental learning HELM. `H-H' and `H-M', for the pairwise HELM based on the HELM and MMD similarity. `UFA' for the UFAN model. `\#' means `number of'. `Val.' for `valid'.}
\end{tabular}
\label{tbl:results}
\end{table}

\section{Conclusions}

In the present paper, several approaches, including a new unsupervised feature alignment network (UFAN), were presented, discussed and compared with the aim of performing unsupervised fault detection on complex industrial systems. 
In a set up were only healthy data can be used for the training, comprising very few operating conditions (data collected over two months only), and only one fault per units, we compared five approaches. 
Two of the approaches relied on single unit training. 
The baseline, trained on two months of data only, proves to be the least performing. 
The extension of this approach with incremental learning can give good results but with two strong limitations: first, for some units the model is not able to learn the new operating conditions, second, the results were only significantly different from the baseline for ratios $r$ above $15\%$, which is not satisfactory for real implementation.

Three approaches rely on pairing the unit of interest with another healthy unit to enhance the training of the one class classifier. We tested two distance measures, HELM as a similarity measure and MMD. We also proposed a new way of performing unsupervised feature alignment with adversarial learning (UFAN).
All these methods are improvements compared to the baseline and the UFAN approach was the best performing. 
These results confirm that the dataset comprises very different units and that finding similar units is challenging. 
Therefore, aligning the features is necessary. 
Aligning is also useful, as it leads to valid model for more than 50\% of all possible pairs. 
As fleet of assets can be of more limited size than in the present case study, this is a strong advantage of the UFAN approach. 

If all the approaches presented here provided an improvement compared to the baseline, it appears that a better selection of the units to pair could provide even better results.
The proposed distances are not good predictors of the model performances, in fact, the correlations are quite low. 
We could demonstrate that pairing any units with UFAN will probably lead to a robust model, yet a limitation is the current impossibility to assess the validity of the models a priori. 
This is of course a major challenge with any unsupervised approach, however this might be mitigated if a better pair selection criteria is identified.

Future work will explore better selection methods to benefit from the excellent results that could be achieved with UFAN if the best pair could have been identified a priori.

\printbibliography

\end{document}